\documentclass[sigconf]{acmart}
\usepackage{multirow}
\usepackage{epstopdf}
\usepackage{subcaption}

\AtBeginDocument{%
  \providecommand\BibTeX{{%
    \normalfont B\kern-0.5em{\scshape i\kern-0.25em b}\kern-0.8em\TeX}}}

\setcopyright{acmcopyright}
\copyrightyear{2018}
\acmYear{2018}
\acmDOI{10.1145/1122445.1122456}
\renewcommand\footnotetextcopyrightpermission[1]{} 

\hypersetup{final}
\begin{document}

\title{Map Enhanced Route Travel Time Prediction using Deep Neural Networks}


\author{Soumi Das}
 \affiliation{
   \institution{IIT Kharagpur}}
 
 \author{Rajath Nandan Kalava}
 \affiliation{
   \institution{IIT Kharagpur}}
 
 \author{Kolli Kiran Kumar}
 \affiliation{
   \institution{IIT Kharagpur}}
 
 \author{Akhil Kandregula}
 \affiliation{
   \institution{IIT Kharagpur}}
 
 \author{Kalpam Suhaas}
 \affiliation{
   \institution{IIT Kharagpur}}
 
 \author{Sourangshu Bhattacharya}
 \affiliation{
   \institution{IIT Kharagpur}}
 
 \author{Niloy Ganguly}
 \affiliation{
   \institution{IIT Kharagpur}}
\begin{abstract}
Travel time estimation is a fundamental problem in transportation science with extensive literature.
The study of these techniques has intensified due to availability of many publicly available large trip datasets.
Recently developed deep learning based models have improved the generality and performance and have focused on estimating times for individual sub-trajectories and aggregating them to predict the travel time of the entire trajectory.
However, these techniques ignore the road network information.
In this work, we propose and study techniques for incorporating road networks along with historical trips' data into travel time prediction. We incorporate both node embeddings as well as road distance into the existing model. Experiments on large real-world benchmark datasets suggest improved performance, especially when the train data is small. As expected, the proposed method performs better than the baseline when there is a larger difference between road distance and Vincenty distance between start and end points.

\end{abstract}

\pagestyle{plain}
\makeatletter
\def\@copyright{\relax}
\makeatother
\if(0)

\ccsdesc[500]{Computer systems organization~Embedded systems}
\ccsdesc[300]{Computer systems organization~Redundancy}
\ccsdesc{Computer systems organization~Robotics}
\ccsdesc[100]{Networks~Network reliability}
\fi
\keywords{Spatio-temporal LSTM, Geo-Convolution, Unmapped Points, Mapped Points, Attribution Error, Node Embedding}


\maketitle

\section{Introduction}
\label{sec:intro}
Travel time estimation \cite{mridha2017link, wang18when} is an important and core research problem in the area of intelligent transportation, with applications in route recommendation, planning and navigation \cite{mridha2017link}, congestion and anomaly detection \cite{chen16congestion}, etc. 
Not surprisingly, the problem has been extensively studied for the past few decades \cite{dharia2003neural,mridha2017link,wang18when}. While a detailed survey is provided in Section \ref{sec:related}, a recent deep learning based approach, called ``Geo-convolution'' \cite{wang18when}, has seen considerable success over its competitors. The key idea of geo-convolution is to extract representations from short segments of trajectories for prediction of their travel times and then stitch them together using multi-task learning. However, the above method does not take the road network into account. In this paper, we study the effectiveness of using the map information explicitly into the prediction process.

We propose to attribute each point in a trajectory to a node in the road network. We utilize two extra pieces of information from the road network: (1)
embeddings of nodes using unsupervised network representation learning techniques e.g. Node2Vec \cite{grover2016node2vec}, and (2) the network/map distance between source and destination points. While the node embeddings are used in the geo-convolution layer, map distance is used for the combined end to end prediction. We evaluate all combinations of proposed techniques on two real world benchmark datasets. We find the additional information indeed improves the overall accuracy of prediction, especially when the number of training trajectories is lower. Additionally, we find that proposed method performs better than the baseline technique in cases where there is a larger difference between actual distance and map distance, confirming the hypothesis that the road network indeed provides additional information for predicting travel times.

\section{Related Work}
\label{sec:related}

Travel time prediction problem has been extensively studied for the past few decades \cite{dharia2003neural,mridha2017link,wang18when}.
While early papers on travel time prediction mostly use static data collection techniques, recent techniques have focused on large amounts of publicly collected GPS traces \cite{wang18when,zhang2018deeptravel}, which have resulted in large training datasets for complex deep learning models. Recent success of Deep Learning in uncovering complex feature interaction patterns has led to its application in link travel time prediction as well. Siripanpornchana et al. \cite{7848343} has used Deep Belief Network with a stack of Restricted Boltzmann Machine to learn the features in unsupervised fashion. Recurrent Neural Networks such as LSTMs are capable of capturing sequential patterns, and have also been adopted to select optimal features automatically \cite{duan16lstm}. Zhang et al. \cite{zhang2018deeptravel} have learned spatio-temporal grid parameters using BiLSTMs and auxiliary supervision for penalizing the deviation from prediction of travel times at intermediate points.  Wang et al. \cite{wang18when} have used geo-convolution layer which captures learned combinations of points in short segments, and a 2-layer stacked LSTM to capture spatial and temporal dependencies among the trajectories. The overall architecture is expected to capture both short term signals from different segments as well as a long term prediction.
\section{Travel Time Prediction}
\label{sec:ttp}
In this section, we describe the problem of trip travel time prediction followed by a recent state-of-the-art baseline approach \cite{wang18when} for predicting trip travel times, without information of the road network. We then describe our proposed modifications (in Section \ref{sec:embconv}) over the baseline approach which uses the information of the road network for predicting the travel times.

\subsection{Problem Definition}
\label{probdefn}

Let $D$ be a given set of trips. A trip $T$ is represented by a sequence of unmapped locations $x_{1} , x_{2} ,...., x_{n}$, the identity of the driver assigned to the trip ($driverID$), the day of the month and week in which the trip occurs ($dateID$, $weekID$), starting time of the trip ($timeID$), distance covered in the trip ($dist$), total time taken for the trip ($time$), and $n-1$ -- $time_{gap}$s and $distance_{gap}$s. Both $time_{gap}$ and $distance_{gap}$ are provided by the difference between time and distance between current location and first location of the trip. Each location $x_i$ is in the form of $\{x_{i1}, x_{i2}\}$, denoting its latitude and longitude. The attribute $timeID$ corresponds to the slot number which is obtained on dividing a day into 1440 timeslots.

Additionally, let $G= (V,E)$ denote the road network where $V$ is the set of all vertices (locations) in the graph. Hence, each element of the set $V$, denoted by $\vec{V_i}$ is in the form of $(v_{i1},v_{i2})$, denoting latitude and longitude of the location $\vec{V_i}$.
An edge set denoted as $E \subset V \times V$, represents the set of road links between locations. 
Given a road network $G$ and a set of unattributed/unmapped trips $D$, we can attribute the locations in the trips to the locations (nodes) in the road network, hence obtaining an attributed/mapped set of trips $\widetilde{D}$.
Hence, each trip $S\in \widetilde{D}$  is characterised by sequence of mapped nodes $y_{1} , y_{2} ,...., y_{n}$, along with $driverID$, $dateID$, $weekID$, $timeID$, $dist$, $time$, $time_{gap}$ and $distance_{gap}$.
The problem of trip travel time prediction thus stands as learning a function $f$ such that it takes a trip $S$ or $T$ and predicts the travel time : $f(T) \in R^+$ (unmapped) or $f(S) \in R^+$ (mapped). We have performed our experiments using the set of mapped trips in all the upcoming methodologies.

\subsection{Node Geo-convolution model}
\label{sec:geoconv}
We consider the baseline model described in \cite{wang18when} to set up the architecture. We term this architecture as \textbf{L-GC}. The authors have used three components in their model - Attribute, Spatio-Temporal, Multi-Task Learning. The Attribute component considers the basic information of the trips viz. driverID, timeID etc, the Spatio-Temporal component tries to learn the spatial and temporal dependencies from the trip nodes, and the Multi-Task learning component relies on the other two components to learn the travel times of a path and also its respective sub-paths.

The main architecture resides in the Spatio-Temporal component. This component is made up of two sub-parts, one being the geo-convolutional neural network which intends to capture the spatial correlation between successive nodes by forming feature maps from the nodes, and the other being the Recurrent Neural Network which captures the temporal correlation in the feature maps.

Each node $x_i$ in the trip sequence is non-linearly mapped into output sequence $loc_i \in R^{16}$ which represents the geographical features of the original node with $W_{loc}$ being the corresponding weight matrix for non linear mapping.
\begin{equation}
loc_i = tanh(W_{loc} . [x_{i1} \cdot x_{i2}])
\end{equation}

The output sequence $loc$ is convolved using a filter of kernel size $k$ with weight matrix $W_{conv}$. This step typically learns the spatial feature of the $i-th$ local path comprising of $k$ nodes.
\begin{equation}
loc_i^{conv} = \sigma_{cnn}( W_{conv} * loc_{i:i+k-1} + b )
\end{equation}

The output of the convolutional layer is appended with the distance of the $i-th$ local path leading to a refined feature map $loc^f$. This resultant map essentially retrieves the spatial dependencies of every local path. In order to capture temporal dependencies among the local paths, Recurrent Neural Networks are added to the output of the geo-convolution layer. Thus, it goes as one of the inputs to the Recurrent Neural Networks aided by the representation vector of the attributes (\textit{attr}) obtained from Attribute component. The Recurrent Neural Network is thus updated with $h_i$ as the hidden state in the following way:
\begin{equation}
h_i = \sigma_{rnn} (W_f . loc_i^f + W_h . h_{i-1} + W_a . attr ) 
\end{equation}

This spatio-temporal sequence of local paths comprising of $h_i$ is converted to a scalar $r_i$ having used fully connected layers. The scalar $r_i$ gives the predicted travel time of local path $i$. In order to predict the travel time of the entire path, the authors have used attention pooling mechanism over the sequence $h_i$. Following this, the local path estimation loss ($L_{local}$) and entire path estimation loss ($L_{global}$) are linearly weighted to train the model.
\begin{equation}
L = \beta . L_{local} + (1 - \beta). L_{global}
\end{equation}

\subsection{Embedding enhanced Geo-convolution model}
\label{sec:embconv}
We develop some modifications over the existing baseline model \cite{wang18when} described in Section \ref{sec:geoconv}. Having used the second variant of trips described in Section \ref{probdefn}, we learn the embeddings of the nodes of graph $G$ using the standard Node2Vec \cite{grover2016node2vec} framework. We use 128-dimensional Node2Vec embeddings for our experiments. The intuition behind using embeddings is that nodes and edges lying close to each other will get similar embedding patterns. This will aid in capturing the spatial dependencies among the roads in a trip. Each node $y_i$ is represented as a 128 dimensional embedded vector. Let $B$ denote the embeddings where each element $b\in B$ is a feature representation of a node. Let $ID$ denote the identities of nodes where each element $i\in ID$ ranges from $\{1,...,n\}$ and n denotes the total number of vertices in the graph G. We rewrite each node $y_i$ in the form $(b_i,i)$ where $b_i$ represents the embedding and $i$ represents the ID of node $y_i$.

Our modifications lie in some proposed changes in the Spatio-Temporal component which stands as the main building block for the entire architecture. Having used the sequence of mapped nodes, we propose two modifications to the baseline model described in Section \ref{sec:geoconv}:

\textbf{Emb-Geo-Conv (E-GC)}: In this method, we map the embedding (Emb) of every node $y_i$ of the trip sequence, denoted by $b_i$ into the output sequence $loc_{emb_i}$. The convolution operation and update of recurrent layer follow the same procedure described in Section \ref{sec:geoconv}. Equation set ~\eqref{eq:1} represent the mapping, convolution operation and recurrent layer update respectively.
\begin{equation}
\begin{gathered}
loc_{emb_i} = tanh(W_{loc} . b_i) \\
loc_{emb_i}^{conv} = \sigma_{cnn}( W_{conv} * loc_{emb_{i:i+k-1}} + b ) \\
h_i = \sigma_{rnn} (W_f . loc_{emb_i}^f + W_h . h_{i-1} + W_a . attr )
\end{gathered}\label{eq:1}
\end{equation}
\textbf{Emb-Node-Geo-Conv (EL-GC)}: We map the concatenation of embedding of $y_i$, denoted by $b_i$ and the mapped node denoted by $y_{i1}$ and $y_{i2}$ into the output sequence $loc_{embcnode_i}$. This is represented using Equation set ~\eqref{eq:2} which follow the convolution operation and recurrent layer update as described in Section \ref{sec:geoconv}
\begin{equation}
\begin{gathered}
loc_{embcnode_i} = tanh(W_{loc} . [b_i \cdot y_{i1} \cdot y_{i2}]) \\
loc_{embcnode_i}^{conv} = \sigma_{cnn}( W_{conv} * loc_{embcnode_{i:i+k-1}} + b ) \\
h_i = \sigma_{rnn} (W_f . loc_{embcnode_i}^f + W_h . h_{i-1} + W_a . attr )
\end{gathered}\label{eq:2}
\end{equation}
\section{Experimental Results}
\label{sec:perf}
In this section, we describe the empirical results comparing existing state-of-the-art \cite{wang18when} and map enhanced techniques described in this paper. We demonstrate our studies with the real world benchmark dataset from the city of Porto and Beijing. In Section \ref{sec:datacreate}, we describe the dataset pre-processing for extraction of trips, acquisition of map from OpenStreetMap \footnote{\url{www.openstreetmap.org}} and the map attribution of the unmapped points. In Section \ref{sec:result}, we compare the baseline technique with new proposed techniques using the metric Mean Absolute Percentage Error(MAPE).

\subsection{Dataset Creation}
\label{sec:datacreate}

\begin{figure}
	\begin{subfigure}{0.49\columnwidth}
		\centering
		\includegraphics[width=\textwidth,height=3cm]{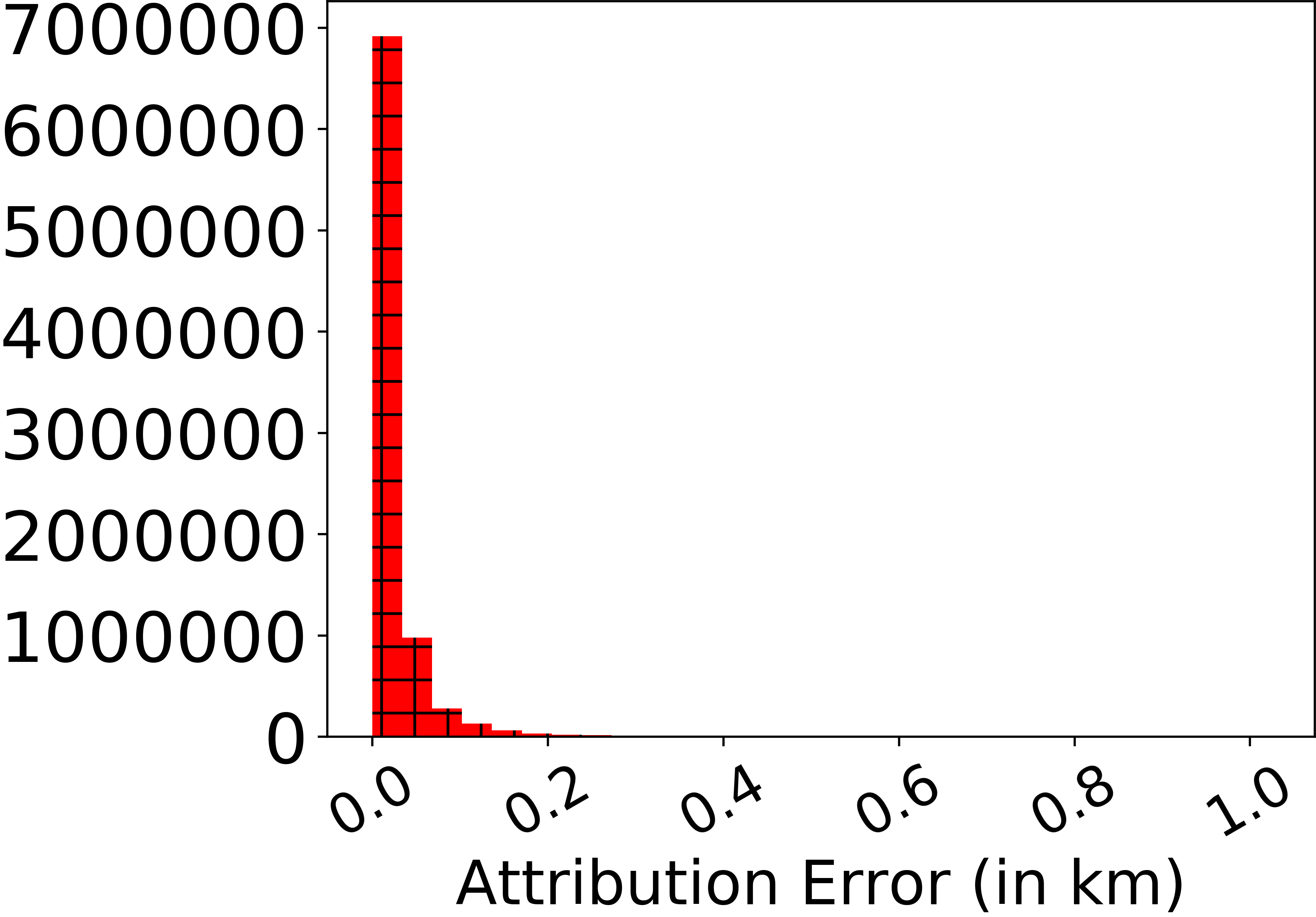}
		\caption{\textbf{Porto Data}}
		\label{fig:attributePorto}
	\end{subfigure}
	\begin{subfigure}{0.49\columnwidth}
		\centering
		\includegraphics[width=\textwidth,height=3cm]{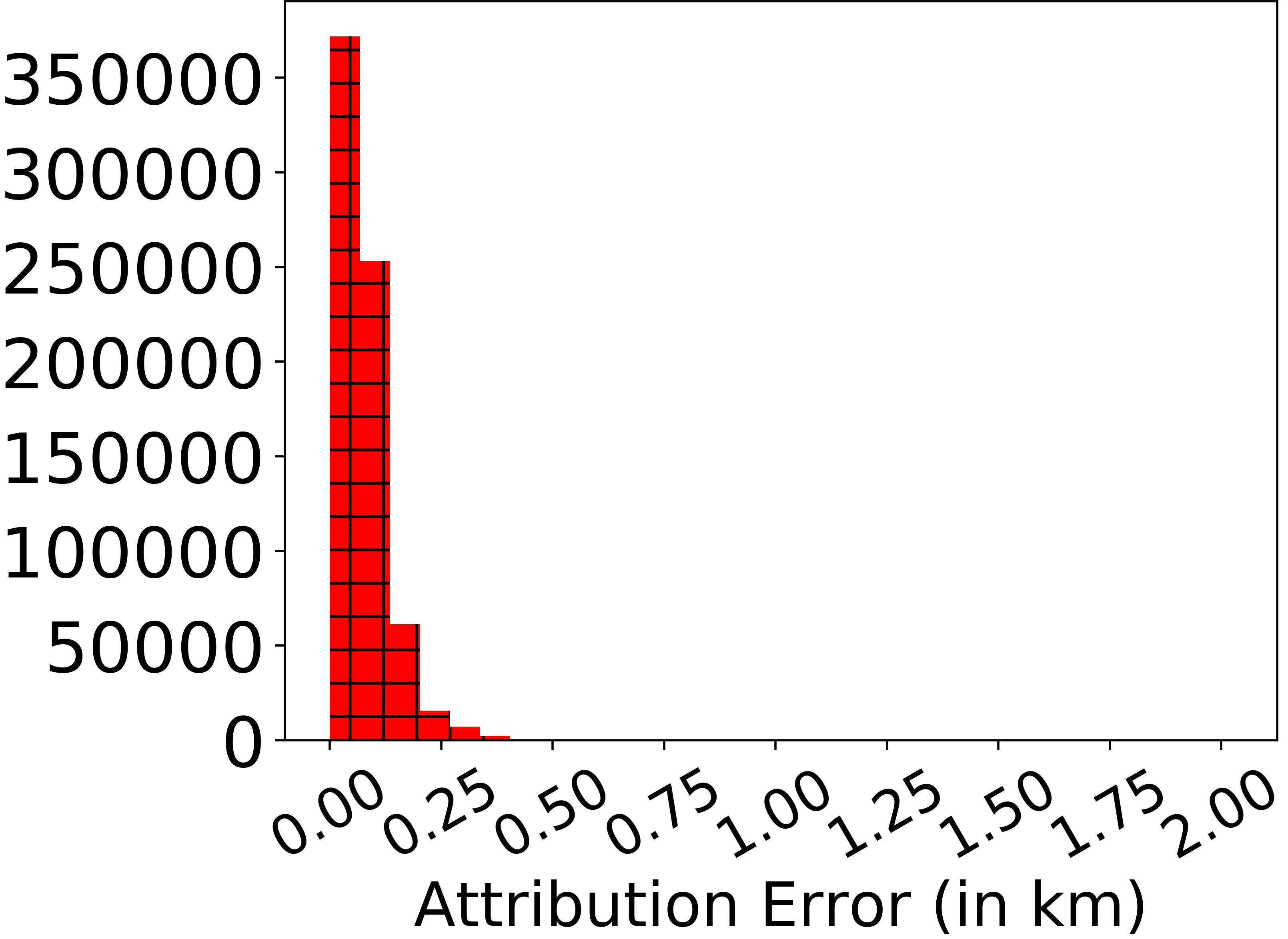}
		\caption{\textbf{Beijing Data}}
		\label{fig:attributeBeijing}
	\end{subfigure}
	\caption{\textbf{Attribution error of trip nodes with frequency of errors on y-axis and error (in km) on x-axis.}}
	\vspace{-5.7mm}
	\label{fig:attr}
\end{figure}

\textbf{Porto GPS Taxi Dataset:} We have used this publicly available dataset for our experiments. Each trip information in this dataset is represented by several attributes among which we used TAXI\_ID, TIMESTAMP and POLYLINE to obtain the trip sequence format described in Section \ref{probdefn}.\\
TAXI\_ID gives the unique number for the taxi driver of that trip, TIMESTAMP denotes the trip's start time and POLYLINE contains a list of GPS coordinates with each pair of coordinates defining 15 seconds of trip.
This dataset consists of a total of 3,26,424 trips, out of which 79\% was used for training, 9\% was used for validation and the remaining was used as the test set.\\
\textbf{Beijing Taxi Dataset:} We have used the sample Beijing dataset provided by the authors in the baseline model \cite{wang18when}. This has been laid out in the format described in Section \ref{probdefn}. The sample dataset consists of a total of 15,772 trips, out of which 68\% was used for training, 23\% was used for validation and the rest was used as the test set.\\
\textbf{OSM Dataset:} Following standard practice in trip travel time prediction e.g. \cite{TOOLE2015162}, we use OpenStreetMap (OSM) data for creating the road network. We use the OSMNX \cite{article2} python package to create the road topology of Porto between longitude range : (-8.835479, -8.285479) and latitude range : (40.893646, 41.443646) and that of Beijing between longitude range : (103.47912, 104.507725) and latitude range : (30.292236, 30.957301). We thus obtain 454842 nodes and 864853 line-strings (connected nodes) from the Porto Map and 179469 nodes and 320649 line-strings from Beijing Map. The justification behind the choice of the above range has been explained later in this section.\\
\textbf{Mapping Points:} We use Locality Sensitive Hashing(LSH) \cite{indyk1998approximate} to map the unmapped location $x_i$ to the closest node $y_i$ in the road network represented by graph G. Owing to the large number of map nodes, we optimize the LSH mapping. We divide the entire area into a number of grids bounded by a max and min pair of (\textit{lat},\textit{lon}). Each OSM node and Porto/Beijing trip data node is mapped into one of the grids. This eventually leads to querying in one of the grids while mapping any trip data node to OSM node, thus reducing the computation time. Figure \ref{fig:attr} demonstrates the distribution of attribution/mapping error on both the datasets, which is measured by the vincenty distance \cite{vincenty1975direct} between each unmapped node in the trip data and its corresponding mapped node from OSM obtained using LSH. As seen from the figure, the attribution/mapping error does not cross 0.2 km (in case of Porto data) and 0.25 km (in case of Beijing data).
\\
\textbf{Formation of trip data:} The original Porto GPS Taxi Dataset encompasses an area between longitude range : (-9.385479, -6.724395) and latitude range : (38.693646, 42.069915). We have selected a region of a specific width (0.55 $\times$ 0.55) where maximum number of trips were covered thus taking the area between longitude range : (-8.835479, -8.285479) and latitude range : (40.893646, 41.443646). Having used TAXI\_ID, TIMESTAMP and POLYLINE features from the dataset and mapping them to OSM nodes using LSH mapping, we modified the existing dataset to create the trip data of mapped nodes in the sequence described in Section \ref{probdefn}. However, in case of Beijing data, we considered the entire region covered in the sample trip data, thus obtaining the map in the previously mentioned range.
\subsection{Results}
\label{sec:result}

In this section, we tabulate the performance of the baseline method (\textbf{L-GC}) beside our proposed modifications (\textbf{E-GC} and \textbf{EL-GC}). We perform the experiments on the two datasets, Porto and Beijing described in Section \ref{sec:datacreate}.

In order to perform the experiments, we have used two types of distance measures:\\
\textbf{Coordinate Distance:} It is found by computing the vincenty distance \cite{vincenty1975direct} between two geographical points (\textit{lat},\textit{lon}). The attributes, $distance_{gap}$ and dist in the data (described in section \ref{probdefn}) have coordinate distance values organized in them.\\
\textbf{Network/Map Distance:} It is computed by finding the shortest path between two points, from the graph obtained using OpenStreetMap (OSM). The points are  the coordinates obtained after mapping the original trip data to OSM (described in Section \ref{sec:datacreate}). Thus, the attributes $distance_{gap}$ and \textit{dist} in the data (described in Section \ref{probdefn}) have the map distance values organized in them.\\
The reason behind using Map Distance is that there will be several pair of nodes which do not have a direct edge between them, but has a series of nodes connecting the pair in the road network. Coordinate distance fails to capture this information of road network. On the contrary, Map distance essentially retrieves the connecting nodes from the road network and returns the shortest path between the individual node in the pair. 

We have reported the Mean Absolute Percentage Error (MAPE) on the test set of both the datasets using the baseline model L-GC and our proposed modifications, E-GC and EL-GC described in Section \ref{sec:geoconv} and \ref{sec:embconv} using the above-mentioned distance measures. Table ~\ref{tab:cumulres} (under Coordinate Distance) shows the comparison of performance metric (MAPE) obtained using these three methodologies and Coordinate Distance as the distance measure.  We observe that both our proposed modifications E-GC and EL-GC perform better in comparison to L-GC (the baseline model \cite{wang18when}) overall, for both the datasets. However, the margin of difference is noted higher in Beijing than that in Porto. We also performed the similar experiment using the Map Distance as the distance measure in Table~\ref{tab:cumulres} (under Map Distance). The proposed modifications performed better than the baseline, with a lower margin.

\begin{table*}[!ptb]
	\footnotesize
	\begin{tabular}{|l|l|l|l|l|l|l|l|l|l|}
		\hline
		\multirow{3}{*}{\textbf{Dataset}} & \multicolumn{9}{c|}{\textbf{Distance Measure}}                                                                                                                                                                                                                                                                                                                                                                                                                     \\ \cline{2-10} 
		& \multicolumn{3}{c|}{\begin{tabular}[c]{@{}c@{}}Map\\ Distance\end{tabular}}                                                                          & \multicolumn{3}{c|}{\begin{tabular}[c]{@{}c@{}}Coordinate\\ Distance\end{tabular}}                                                                   & \multicolumn{3}{c|}{\begin{tabular}[c]{@{}c@{}}Map \&\\ Coordinate\\ Distance\end{tabular}}                                                          \\ \cline{2-10} 
		& \begin{tabular}[c]{@{}l@{}}L-GC\end{tabular} & \begin{tabular}[c]{@{}l@{}}E-GC\end{tabular} & \begin{tabular}[c]{@{}l@{}}EL-GC\end{tabular} & \begin{tabular}[c]{@{}l@{}}L-GC\end{tabular} & \begin{tabular}[c]{@{}l@{}}E-GC\end{tabular} & \begin{tabular}[c]{@{}l@{}}EL-GC\end{tabular} & \begin{tabular}[c]{@{}l@{}}L-GC\end{tabular} & \begin{tabular}[c]{@{}l@{}}E-GC\end{tabular} & \begin{tabular}[c]{@{}l@{}}EL-GC\end{tabular} \\ \hline
		Beijing                           & 31.08                                           & 28.54                                           & \textbf{27.32}                                            & 30.66                                           & 29.51                                           & \textbf{26.23}                                            & 31.05                                           & 28.32                                                & \textbf{26.01}                                                 \\ \hline
		Porto                             & 14.90                                           & 14.75                                           & \textbf{14.67}                                            & 14.67                                           & 14.45                                           & \textbf{14.42}                                           & 14.65                                                & 14.61                                                 & \textbf{14.53}                                                  \\ \hline
	\end{tabular}
	\caption{Comparison of Baseline (L-GC) against the modifications (E-GC and EL-GC) on the basis of Mean Absolute Percentage Error (MAPE \%) using Coordinate Distance, Map Distance and both combined.}
	\vspace{-7mm}
	\label{tab:cumulres}
\end{table*}

\begin{figure}
	\begin{subfigure}{0.49\columnwidth}
		\centering
		\includegraphics[width=\textwidth,height=3cm]{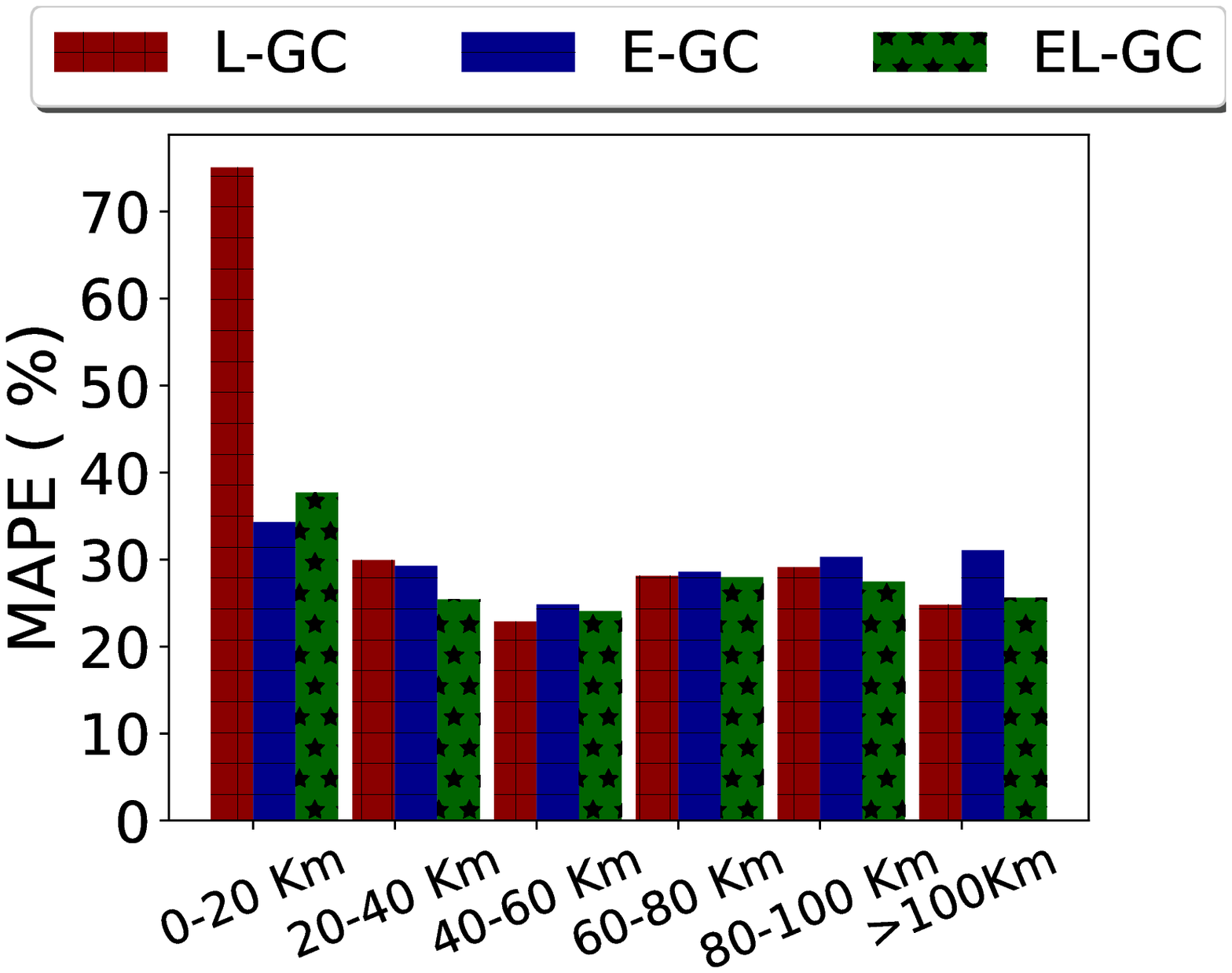}
		\caption{\textbf{Coordinate Distance}}
		\label{fig:attrmapeCoord}
	\end{subfigure}
	\begin{subfigure}{0.49\columnwidth}
		\centering
		\includegraphics[width=\textwidth,height=3cm]{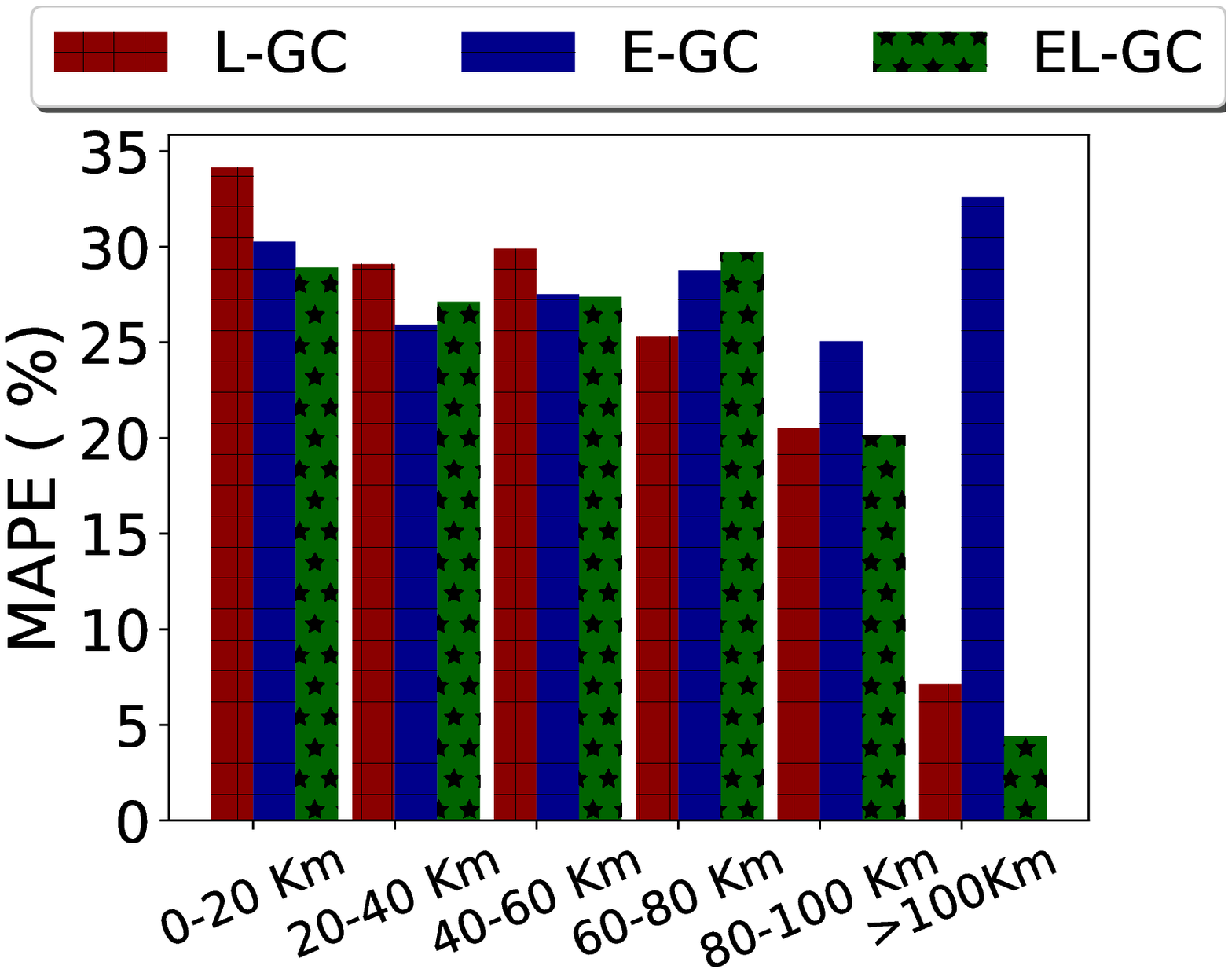}
		\caption{\textbf{Map Distance}}
		\label{fig:attrmapeMap}
	\end{subfigure}
	\caption{\textbf{Distribution of MAPE(y-axis) across L-GC, E-GC and EL-GC over varying Coordinate and Map distance buckets (x-axis) on Beijing Dataset.}}
	\vspace{-5mm}
	\label{fig:attrmape}
\end{figure}

We also show the variation of MAPE across different distance buckets for Beijing dataset. Figure~\ref{fig:attrmapeCoord} shows the variation across Coordinate distance bucket while Figure~\ref{fig:attrmapeMap} shows the variation across Map distance bucket. It is observed that L-GC has a very high MAPE for the shorter path trips (with Coordinate distance), which eventually keeps on decreasing with longer trips. Alongside, E-GC and EL-GC perform better across all length paths. For paths with Map distance, L-GC and EL-GC show similar performance while E-GC degrades in performance for longer path trips. However, in overall EL-GC performs the best across all length trips of both Map and Coordinate distance.

\begin{figure}
	\includegraphics[width=0.6\columnwidth,height=3cm]{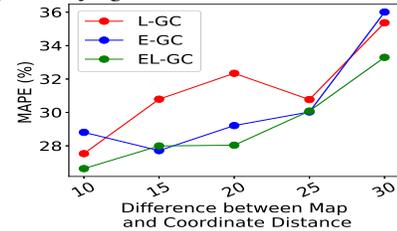}
	\caption{\textbf{Distribution of MAPE(y-axis) across L-GC, E-GC and EL-GC over the varying difference between Coordinate and Map distance buckets (x-axis) on Beijing Dataset.}}
	\vspace{-5mm}
	\label{fig:mapcoorddiff}
\end{figure}

Following the same set of experiments, we modified on the grounds of using distance metric. We concatenated both the Map Distance and Coordinate Distance to the output of the convolution layer ($loc_i^{conv}$). Table~\ref{tab:cumulres} (under Map \& Coordinate Distance) shows the comparison of the baseline method L-GC with that of the two modifications E-GC and EL-GC. We observe that both the modifications E-GC and EL-GC outperform L-GC. We also plot the MAPE across varying buckets holding the difference between Map Distance and Coordinate Distance. The assumption that holds in coordinate distance, that all nodes are directly connected is violated by map distance which includes the map information to find the shortest distance between two nodes (which can be connected by a series of nodes in map). It can be clearly seen from Figure~\ref{fig:mapcoorddiff} that although at a lower difference value in distance (on x-axis), L-GC starts at a lower MAPE, it keeps increasing as the difference gets higher. On the other hand, E-GC and EL-GC remain on the lower side as the difference in distance increases. The inclusion of embeddings along with the nodes help in a better interpretation of spatial dependencies among the nodes thus aiding in better prediction of the trips with large difference between map and coordinate distance. This suggests that our proposed modifications perform better for trips which have the subsequent nodes connected by a series of nodes in the road network (and not directly connected).

\section{Conclusion}
In this paper, we study the problem of travel time prediction of any given path. We build our modifications on an existing state-of-the-art method \cite{wang18when} based on deep neural networks. We incorporate Map Distance and node embeddings in the network which prove to perform better compared to the baseline method. The information from the road network aided by the embeddings which essentially captures the vicinity of nodes around each other, account for the better performance of the modifications, thus confirming our hypothesis.
\bibliographystyle{ACM-Reference-Format}
\bibliography{References/ref}
\end{document}